\pdfoutput=1

\documentclass[11pt]{article}

\usepackage[]{ACL2023}

\definecolor{c1}{cmyk}{0,0.6175,0.8848,0.1490}
\definecolor{c2}{cmyk}{0.1127,0.6690,0,0.4431}
\definecolor{c3}{cmyk}{0.3081,0,0.7209,0.3255}
\definecolor{c4}{cmyk}{0.6765,0.2017,0,0.0667}
\definecolor{c5}{cmyk}{0,0.8765,0.7099,0.3647}

\definecolor{gg}{HTML}{0F9D58}
\definecolor{rr}{HTML}{DB4437}
\definecolor{bb}{HTML}{4285F4}

\usepackage{hyperref}
\usepackage{nameref}

\usepackage{times}
\usepackage{latexsym}
\usepackage{graphicx}
\usepackage{algorithm, algorithmic}
\usepackage{amsmath}
\usepackage{booktabs}
\usepackage{tabularx}
\usepackage{colortbl}
\usepackage{hhline}
\usepackage{amssymb} 
\usepackage{array} 
\usepackage{multirow} 
\usepackage{enumitem}

\usepackage{graphicx}
\usepackage{booktabs}
\usepackage{amssymb} 
\usepackage{xcolor}  
\usepackage[T1]{fontenc}

\usepackage{algorithm}
\usepackage{algorithmic}
\usepackage{array}
\usepackage{amsmath}
\usepackage{amsfonts}
\usepackage{booktabs} 
\usepackage{arydshln} 
\usepackage{multicol}
\usepackage{multirow}
\usepackage[table]{xcolor}

\usepackage[most]{tcolorbox}

\newtcolorbox{prompt}[1]{
    enhanced,
    drop shadow=black!5!white,
    left=4mm,
    right=4mm,
    top=2mm,
    bottom=2mm,
    boxsep=0mm,
    rounded corners,
    title=#1,
    fontupper=\footnotesize\linespread{0.9}\fontfamily{lmr}\selectfont,
    }

\usepackage[utf8]{inputenc}

\usepackage{microtype}

\usepackage{inconsolata}

\title{"\textit{Excuse me, may I say something...}" CoLabScience, A Proactive AI Assistant for Biomedical Discovery and LLM-Expert Collaborations}

\author{
\textbf{Yang Wu}\textsuperscript{$\clubsuit$} \quad
\textbf{Jinhong Yu}\textsuperscript{$\clubsuit$} \quad
\textbf{Jingwei Xiong}\textsuperscript{$\heartsuit$} \\
\textbf{Zhimin Tao}\textsuperscript{$\diamondsuit$} \quad
\textbf{Xiaozhong Liu}\textsuperscript{$\clubsuit$}\thanks{\, Corresponding author.} \\
\textsuperscript{$\clubsuit$}Worcester Polytechnic Institute, Worcester, MA, USA \\
\textsuperscript{$\heartsuit$}University of California, Davis, CA, USA \\
\textsuperscript{$\diamondsuit$}Jiangsu University, Zhenjiang, Jiangsu, China \\
\texttt{\{ywu19, jyu7, xliu14\}@wpi.edu} \\
\texttt{jwxxiong@ucdavis.edu}\quad
\texttt{jsutao@ujs.edu.cn}
}

\begin{document}
\maketitle

\begin{abstract}
The integration of Large Language Models (LLMs) into scientific workflows presents exciting opportunities to accelerate biomedical discovery. However, the reactive nature of LLMs, which respond only when prompted, limits their effectiveness in collaborative settings that demand foresight and autonomous engagement. In this study, we introduce CoLabScience, a proactive LLM assistant designed to enhance biomedical collaboration between AI systems and human experts through timely, context-aware interventions. At the core of our method is PULI (\textit{Positive-Unlabeled Learning-to-Intervene}), a novel framework trained with a reinforcement learning objective to determine when and how to intervene in streaming scientific discussions, by leveraging the team's project proposal and long- and short-term conversational memory. To support this work, we introduce BSDD (\textit{Biomedical Streaming Dialogue Dataset}), a new benchmark of simulated research discussion dialogues with intervention points derived from PubMed articles. Experimental results show that PULI significantly outperforms existing baselines in both intervention precision and collaborative task utility, highlighting the potential of proactive LLMs as intelligent scientific assistants.\footnote{https://github.com/YANGWU001/CoLabScience.}
\end{abstract}

\section{Introduction}

\begin{figure}[t]
\centering 
\includegraphics[width=\linewidth]{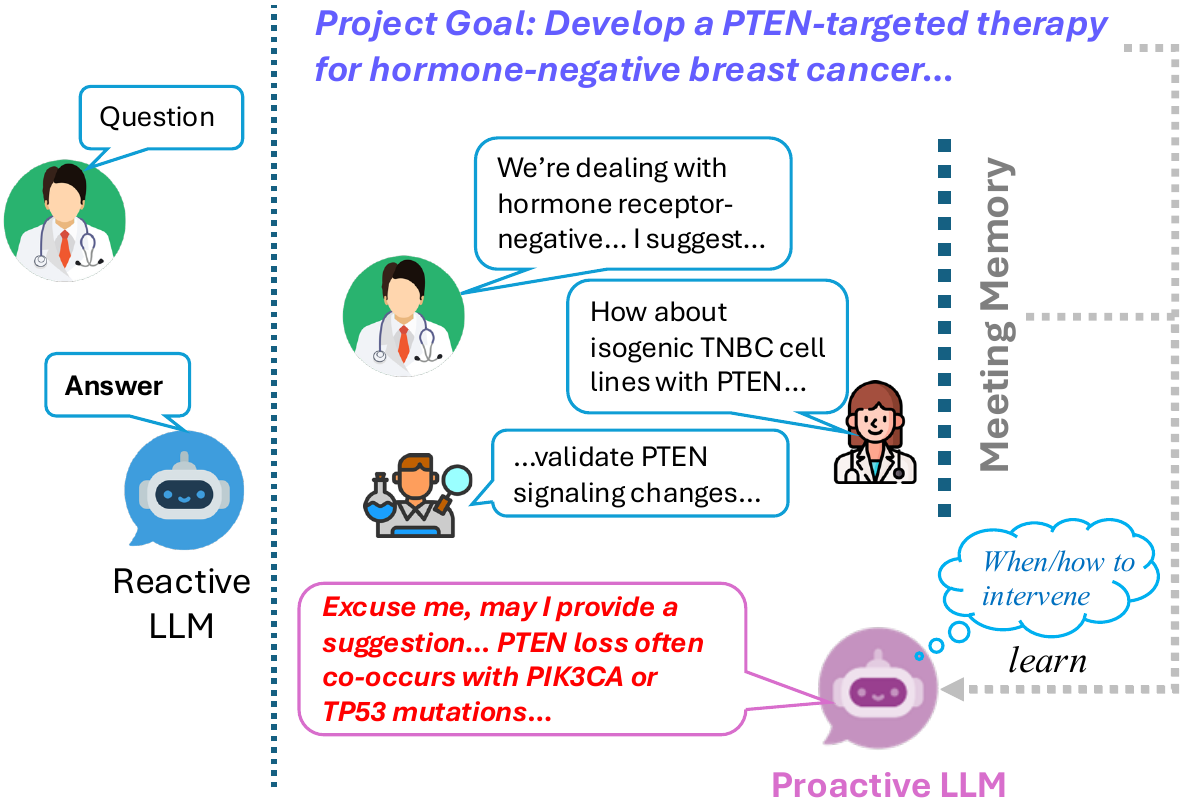}
\caption{Comparison of Reactive and Proactive LLMs in Biomedical Collaboration. Traditional reactive LLMs (left) respond only after being prompted, while proactive LLMs (right) monitor ongoing discussions, identify opportunities to contribute domain-relevant insights, and intervene with timely and context-aware suggestions without explicit prompting.
} 
\label{fig:toy_example}
\end{figure}
Recent developments in large language models (LLMs) have fostered advancements in scientific research, enabling accelerated discovery in biomedical fields \citep{luo2022biogpt,ma2024mol,jin2025stella}. In particular, existing work has explored their potential across tasks such as drug repurposing, disease diagnosis, and clinical question answering \citep{qi2024large,zhao2023survey,lu2024large}. Despite these successes, current models primarily function in a reactive paradigm \citep{liao2023proactive,lu2024large,yao2025elevating}, responding solely upon explicit prompts from researchers. This interaction mode significantly restricts their effectiveness in collaborative settings, where the absence of proactive interventions can lead to missed critical insights and opportunities \citep{yang2025socialmind,wu2024knowledge}. In response to these limitations, we propose that LLMs supporting biomedical research should evolve toward proactive engagement: continuously tracking ongoing discussions, understanding emerging contexts, and autonomously identifying appropriate moments for contribution—integrating into the team as an active team member rather than remaining a passive tool. For instance, as illustrated in Figure 1, a traditional reactive LLM \citep{hassouna2025llm,zhou2026toward} responds passively only after being explicitly prompted by the Clinical Physician, whereas a proactive LLM tracks the discussion, identifies timely opportunities to contribute PTEN-relevant insights, and initiates suggestions that facilitate scientific progress without waiting for direct queries.

Motivated by the need for proactive engagement, we present CoLabScience, a novel AI assistant that transforms LLMs from reactive tools to proactive collaborators in biomedical research. At its core lies PULI (\textit{Positive-Unlabeled Learning-to-Intervene}), a framework trained with reinforcement learning to determine when and how to intervene during scientific discussions. To train this model, we constructed BSDD (\textit{Biomedical Streaming Dialogue Dataset}), a collection of simulated scientific dialogues characterized by multiple research roles (e.g., Pharmacologist, Clinical Physician), generated by LLMs with content grounded in PubMed literature \citep{sayers2024database}. To ensure the reliability of LLM-derived labels and mitigate hallucination risk \citep{huang2025survey, sriramanan2024llm}, we adopt a sparse labeling strategy in which only the most valuable intervention points are labeled as positive, while all others remain unlabeled \citep{kiryo2017positive,wu2023community}.

By leveraging CoLabScience's coordinator to identify reliable negative interventions from the positive-unlabeled (PU) data, we employ a two-tier approach: training a small \textbf{\textit{Observer}} LLM with Group Relative Policy Optimization (GRPO) \citep{shao2024deepseekmath} to determine when to intervene and fine-tuning a large-scale \textbf{\textit{Presenter}} LLM with supervised learning to generate appropriate intervention content. This architecture enables real-time dialogue monitoring via the efficient Observer model, invoking the computationally expensive Presenter model only when an intervention is needed. Reward signals from both models are integrated to train the reinforcement learning coordinator, enabling an end-to-end training loop. By incorporating the project proposal (i.e., project goals, datasets, and background knowledge) and dual-scale conversational memory, in which long-term memory retains critical prior insights and short-term memory captures the evolving conversational context \citep{hatalis2023memory,zhong2024memorybank}, CoLabScience proactively delivers scientifically grounded interventions without requiring explicit prompting.

Our contributions in this paper are threefold and can be summarized as follows:

$\bullet$ We propose CoLabScience, a proactive LLM assistant that supports efficient biomedical research collaboration through context-aware interventions. Unlike reactive LLMs, CoLabScience autonomously determines when and how to intervene during ongoing research discussions by leveraging project context and conversational history.\par
$\bullet$ We introduce BSDD, a new open benchmark consisting of simulated biomedical research dialogues grounded in PubMed articles and annotated with proactive intervention labels. BSDD provides a valuable resource for training and evaluating future proactive scientific assistants, advancing research in this emerging area.\par
$\bullet$ We empirically validate CoLabScience's effectiveness through both simulation-based evaluation and human evaluation. The results demonstrate the model’s strong generalization ability and robustness across a range of LLM backbones.

\section{Related Work} 

\paragraph{Large Language Models as Scientific Assistants}
Recent advances in LLMs have shown promise in biomedical research through protein structure prediction, antibiotic discovery, and drug repurposing \citep{wong2024discovery,zambaldi2024novo,jumper2021highly,swanson2024virtual,gottweis2025towards,yao2025intelligent}. Beyond the biomedical domain, systems like Agent Laboratory and AI Scientist \citep{schmidgall2025agent,lu2024ai} automate research pipelines from hypothesis generation to reporting, while Agentic Reasoning \citep{wu2025agentic} enhances multi-step reasoning. Complementing these systems, RECODE-H~\citep{miao2025recode} benchmarks human-agent collaboration through multi-turn code development. However, these approaches are largely reactive, whereas CoLabScience enables context-aware, timely interventions during ongoing scientific discussions.

\paragraph{Proactive Capabilities in Large Language Models}
Recent work explores LLM proactivity through structured prompting, including initiative-taking in collaboration \citep{zhang2024proagent,lu2024proactive,wu2025teaching}, clarification-seeking behaviors \citep{zhang2024ask,qian2024tell,liu2024compeer,panginteractive,li2024mediq,wu2024rose}, and requesting user support in complex tasks \citep{wu2024need}. VideoLLM-Online \citep{chen2024videollm} extends this to multimodal streaming, training models to determine optimal narration timing. However, these approaches rely on hand-crafted prompts and fixed logic, limiting adaptive intervention. In contrast, we introduce a trainable reinforcement learning mechanism that enables context-aware, timing-sensitive proactive decision-making.

\section{Methodology}
\label{method}

\subsection{Preliminary Definitions}

We formalize the proactive intervention task based on multi-round scientific dialogues, where each round corresponds to a single utterance from one team member. Let $\mathcal{D} = \{ D^1, D^2, \ldots, D^M \}$ denote a collection of independent multi-round dialogues. Each dialogue $D^i = \{ d^i_1, d^i_2, \ldots, d^i_{N_i} \}$ consists of $N_i$ rounds generated from a fixed project proposal $C^i$, which defines the research goal, background knowledge, and relevant datasets. We aggregate all dialogue rounds into a positive-unlabeled (PU) intervention training set:
\[
\{d_1, d_2, \ldots, d_N\},
\]
where $N = \sum_{i=1}^{M} N_i$ is the total number of candidate rounds. The PU training set contains $u$ unlabeled rounds $U = \{d_1, \ldots, d_u\}$ with unknown intervention necessity, and $(N-u)$ labeled positive rounds $P = \{d_{u+1}, \ldots, d_N\}$.

Based on the PU dataset, our task is to jointly learn (1) when to intervene, using a Observer $\mathcal{H}_\phi$, and (2) how to intervene, using a Presenter $\mathcal{G}_\psi$. To facilitate the training of both LLMs, we introduce a reinforcement learning framework where a coordinator model $\mathcal{F}_\theta$ identifies potential positive and negative samples from the unlabeled data. To enhance understanding and clarity, a summary of notations is provided in Table~\ref{tab:notations}.


\begin{table}[h]
\centering
\scriptsize 
\setlength{\tabcolsep}{3pt} 
\renewcommand{\arraystretch}{0.95} 
\begin{tabular}{p{0.34\columnwidth} p{0.60\columnwidth}}
\toprule
\textbf{Notations} & \textbf{Descriptions} \\
\midrule
$\mathcal{D}=\{ D^1,\ldots,D^M \}$ & Set of $M$ multi-round dialogues. \\
$D^i=\{d^i_1,\ldots,d^i_{N_i}\}$ & $i$-th dialogue with $N_i$ rounds. \\
$C^i$ & Project proposal context of $D^i$ (goal, background, datasets). \\
$U=\{d_1,\ldots,d_u\}$ & Unlabeled intervention rounds. \\
$P=\{d_{u+1},\ldots,d_N\}$ & Positive intervention rounds. \\
$d_n$ & $n$-th unlabeled intervention round. \\
$C(d_n)$ & Project proposal associated with the original dialogue of $d_n$. \\
$\mathcal{M}^L(d_n)$ & Long-term memory of $d_n$. \\
$\mathcal{M}^S(d_n)$ & Short-term memory of $d_n$. \\
$\mathcal{M}(d_n)$ & Overall contextualized memory. \\
$\mathcal{F}_\theta$ & Coordinator model. \\
$\mathcal{H}_\phi$ & Observer LLM for intervention timing. \\
$\mathcal{G}_\psi$ & Presenter LLM for intervention responses. \\
\bottomrule
\end{tabular}
\caption{Notations.}
\label{tab:notations}
\end{table}

\begin{figure}[t]
\centering 
\includegraphics[width=\linewidth]{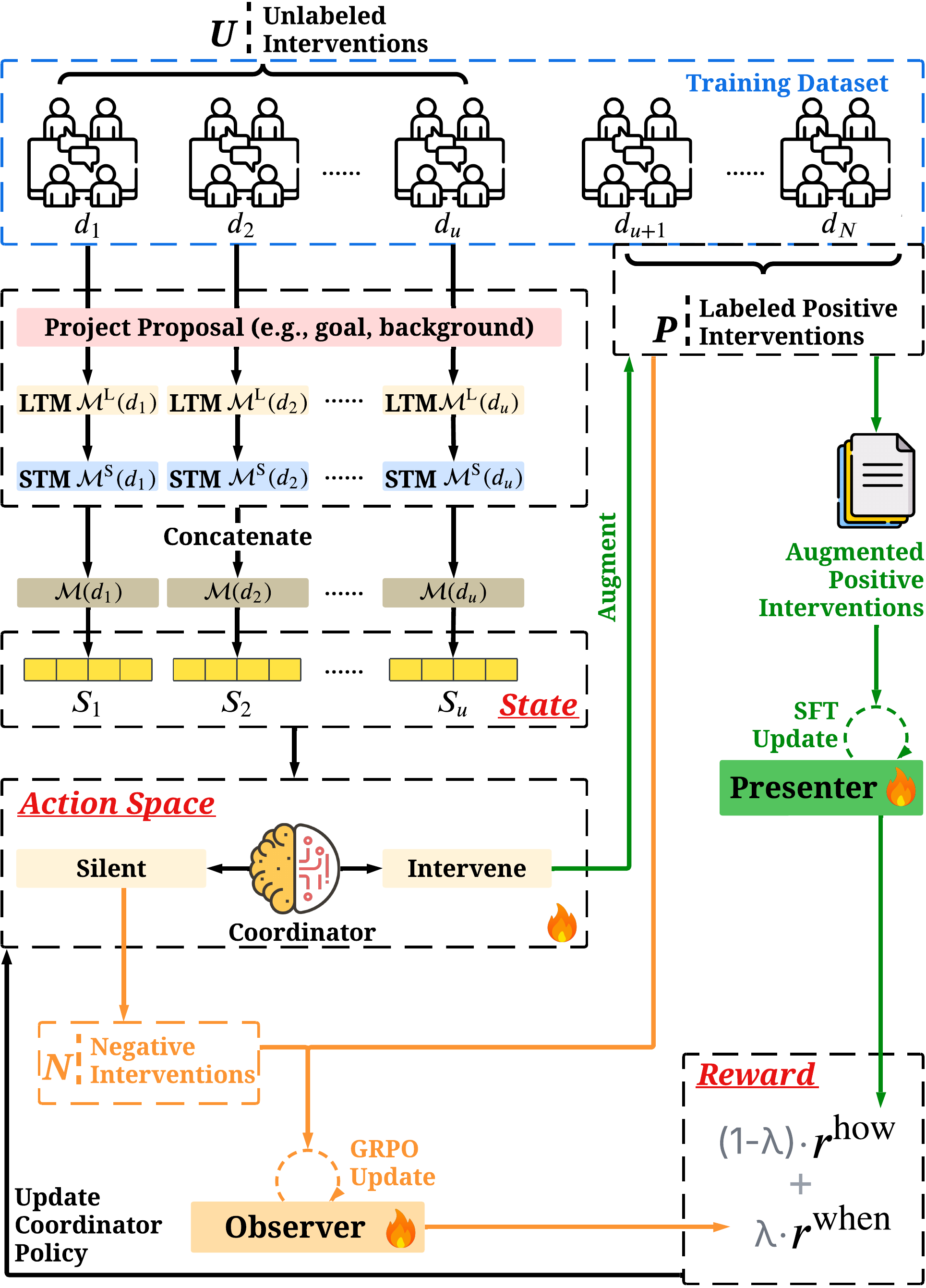}
\caption{Illustration of PULI framework. The coordinator decides whether to intervene or remain silent for each unlabeled dialogue round. Silent rounds are used as negative samples to update the Observer through GRPO training to learn intervention timing, while intervention rounds augment positive data to refine the Presenter to generate appropriate intervention content. The Observer and Presenter collaboratively provide rewards to optimize the coordinator in an end-to-end training process.} 
\label{fig:framework}
\end{figure}
\subsection{PULI Mechanism}

In this section, we introduce the PULI mechanism, which jointly learns when and how to intervene in multi-round scientific dialogues using a positive-unlabeled dataset. An overview of the framework is shown in Figure~\ref{fig:framework}.

\subsubsection{Multi-Round Dialogue State}
\label{method:memory}
For each unlabeled intervention round $d_n \in U$ at local dialogue step $t$, we construct a contextualized memory $\mathcal{M}_t(d_n)$ comprising three components:
(1) the project proposal $C(d_n)$, specifying the research goal, background, and data information;
(2) a short-term memory $\mathcal{M}_t^S(d_n)$, capturing the current utterance and its two most recent predecessors;
(3) a long-term memory $\mathcal{M}_t^L(d_n)$, summarizing accumulated meeting insights up to step $t$. The memory construction is defined as:
\begin{equation}
\small
\begin{aligned}
\mathcal{M}_t^S(d_n) &= \{d_n^{t-2}, d_n^{t-1}, d_n^t\}, \\
\mathcal{M}_t^L(d_n) &= 
\begin{cases}
\emptyset, & \text{if } t = 0, \\
\Gamma\left(\mathcal{M}_{t-1}^L(d_n) \cup \mathcal{M}_{t-1}^S(d_n)\right), & \text{if } t > 0, \\
\end{cases} \\
\mathcal{M}_t(d_n)   &= [C(d_n), \mathcal{M}_t^S(d_n), \mathcal{M}_t^L(d_n)].
\end{aligned}
\label{eq:memory}
\end{equation}

For initial steps where $t < 2$, we omit unavailable indices in $\mathcal{M}_t^S(d_n)$, e.g., $\mathcal{M}_0^S(d_n) = \{d_n^0\}$ and $\mathcal{M}_1^S(d_n) = \{d_n^0, d_n^1\}$. This design ensures that the short-term memory is well-formed at all time steps, while the long-term memory recursively compresses earlier rounds via the LLM summarizer $\Gamma(\cdot)$ to prevent excessive memory accumulation.

To obtain the reinforcement learning state, we process the memory input $\mathcal{M}_t(d_n)$ through both the Observer $\mathcal{H}\phi$ and the Presenter $\mathcal{G}_\psi$, and extract their final hidden representations to construct the state embedding $S_n$:

\begin{equation}
\small
S_n = \text{Concat}\left(\Psi_{\mathcal{H}_\phi}\left(\mathcal{M}_t(d_n)\right), \Omega\left(\Psi_{\mathcal{G}_\psi}\left(\mathcal{M}_t(d_n)\right)\right)\right),
\label{eq:embedding}
\end{equation}

where $\Psi_{\mathcal{H}_\phi}(\cdot)$ and $\Psi_{\mathcal{G}_\psi}(\cdot)$ denote the last hidden layer representations of $\mathcal{H}_\phi$ and $\mathcal{G}_\psi$, respectively, and $\text{Concat}(\cdot, \cdot)$ denotes the concatenation operator. To ensure dimensional consistency between the last hidden layer representations of the Observer and Presenter, we introduce a learnable linear projector $\Omega(\cdot)$ that projects $\Psi_{\mathcal{G}_\psi}\left(\mathcal{M}_t(d_n)\right)$ into the same dimension as the $\Psi_{\mathcal{H}_\phi}\left(\mathcal{M}_t(d_n)\right)$. The projector is jointly optimized with the coordinator policy $\mathcal{F}_\theta$ to determine whether to intervene at round $d_n$.

\subsubsection{Silent or Intervene?}

To identify potential intervene (positive) and silent (negative) samples from the PU data, the coordinator in our PULI framework formulates a binary decision-making problem. Specifically, for each unlabeled candidate round \( d_n \), the coordinator observes the dialogue state \( S_n \) and selects an action \( a_n \in \{0,1\} \), where \( a_n = 1 \) indicates choosing to intervene, and \( a_n = 0 \) corresponds to remaining silent. Formally, we implement the coordinator \( \mathcal{F}_\theta \) as a multilayer perceptron (MLP), which takes the state \( S_n \) as input and outputs an intervention probability \( \mathcal{F}_\theta(S_n) \) ranging from 0 to 1. The MLP uses ReLU activations in hidden layers and a sigmoid function in the output layer to produce probabilistic decisions. We adopt a fixed decision threshold of 0.5 to map \( \mathcal{F}_\theta(S_n) \) to a binary action: \(a_n=1\) if \( \mathcal{F}_\theta(S_n)\ge 0.5 \), and \(a_n=0\) otherwise.

The coordinator's policy function is defined as:
\begin{equation}
\small
\begin{aligned}
\pi_\theta(S_n, a_n) 
&= P_\theta(a_n \mid S_n) \\
&= a_n \cdot \mathcal{F}_\theta(S_n) + (1 - a_n) \cdot (1 - \mathcal{F}_\theta(S_n)),
\end{aligned}
\end{equation}
where \( \pi_\theta(S_n, a_n) \) denotes the probability of selecting action \( a_n \) given state \( S_n \). This design enables the coordinator \( \mathcal{F}_\theta \) to learn to identify reliable negative examples (i.e., rounds to remain silent) from unlabeled intervention candidates, supporting effective policy refinement.

\subsubsection{Learning-to-Intervene Reward}
\label{method:reward}

At each global training epoch \( T \) of PULI, after the coordinator selects a set of negative interventions \( \mathcal{N} \), these negative samples are combined with the labeled positive interventions \( \mathcal{P} \) from the training dataset to construct a binary supervision signal. The Observer is then trained using GRPO \citep{shao2024deepseekmath}, where it is rewarded with 1 for correctly identifying a sample's intervention label and 0 otherwise. The model’s performance is evaluated on a held-out validation set, and we denote the validation accuracy at epoch \( T \) as \( z^T \).

To effectively encourage learning, we compare the current Observer's performance against the best historical performance. Specifically, we define the reward for intervention timing, \( r^{\text{when}} \), as:
\begin{equation}
r^{\text{when}} = z^T - x^T,
\end{equation}
where \( x^T = \max(z^0, z^1, \ldots, z^{T-1}) \) represents the best validation accuracy achieved in previous epochs, and \( z^0 \) corresponds to the Observer trained at initialization by treating all unlabeled instances as negative.

In parallel, the selected positive samples are used to augment the labeled positive intervention set, resulting in an expanded positive set \( \mathcal{P}' \). This set is used to train the Presenter via supervised fine-tuning. To evaluate the quality of generated interventions, we compute the ROUGE-1 score of the LLM on held-out validation examples. Let the ROUGE-1 score at epoch \( T \) be denoted as \( l^T \). The reward for intervention content quality, \( r^{\text{how}} \), is defined as
\begin{equation}
r^{\text{how}} =  l^T - h^T ,
\end{equation}
where \( h^T = \max(l^0, l^1, \ldots, l^{T-1}) \) is the highest validation ROUGE-1 score observed up to epoch \( T-1 \).

Finally, the total reward \( r_{\text{total}} \) used to update the coordinator is a weighted combination of the two components:
\begin{equation}
\label{reward_function_sum}
r_{\text{total}} = \lambda \cdot r^{\text{when}} + (1 - \lambda) \cdot r^{\text{how}},
\end{equation}
where \( \lambda \in [0,1] \) balances the importance between \textit{when} to intervene and \textit{how} to intervene. In all experiments, we set the default value of \( \lambda = 0.6 \).

\subsection{Model Training}

\paragraph{Coordinator Optimization}
Inspired by~\citep{luo2021pulns}, we train the coordinator $\mathcal{F}_\theta$ using a reinforcement learning objective with the reward function defined in Equation~\ref{reward_function_sum}. At each training epoch $T$, the coordinator observes the dialogue state $S_n$ for each unlabeled round $d_n \in \mathcal{U}$ and samples an action $a_n$ from its policy $\pi_\theta(S_n)$. The resulting action sequence is evaluated using $r_{\text{total}}$, which integrates improvements in both intervention timing and content quality. Given the trajectory \( \tau = \{(S_n, a_n)\}_{n=1}^u \) in epoch \( T \), the coordinator's objective is to maximize the expected reward:
\begin{equation}
J(\theta) = \mathbb{E}_{\pi_\theta} \left[ r_{\text{total}}^T \right].
\end{equation}
We apply the REINFORCE algorithm to compute the policy gradient:
\begin{equation}
\nabla_\theta J(\theta) \approx \sum_{n=1}^{u} r_{\text{total}}^T \cdot \nabla_\theta \log \pi_\theta(S_n, a_n),
\end{equation}
and update the coordinator parameters with learning rate $\eta$:
\begin{equation}
\theta \leftarrow \theta + \eta \sum_{n=1}^{u} r_{\text{total}}^T \cdot \nabla_\theta \log \pi_\theta(S_n, a_n),
\end{equation}

\paragraph{End-to-End Training}
The overall training procedure is performed in an end-to-end iterative loop that jointly updates the coordinator, Observer, and Presenter. Initially, the Observer and Presenter are independently pre-trained by treating all unlabeled rounds in \( \mathcal{U} \) as negative. The coordinator is randomly initialized.

At each epoch, the coordinator observes each unlabeled dialogue round \( d_n \), encodes its state \( S_n \) using the last hidden layer representations from the Observer and Presenter, and samples an action \( a_n \). Rounds selected for intervention (\( a_n = 1 \)) are used to augment the positive set for updating the Presenter, while rounds with \( a_n = 0 \) are treated as negatives to refine the Observer. After training both LLMs, the resulting changes in Observer accuracy and Presenter ROUGE-1 scores are used to compute the total reward \( r_{\text{total}}^T \) and update the coordinator via policy gradient. This iterative process continues for a fixed number of epochs, leading to an end-to-end training framework to optimize intervention strategies.

\begin{figure}[t]
\centering 
\includegraphics[width=\linewidth]{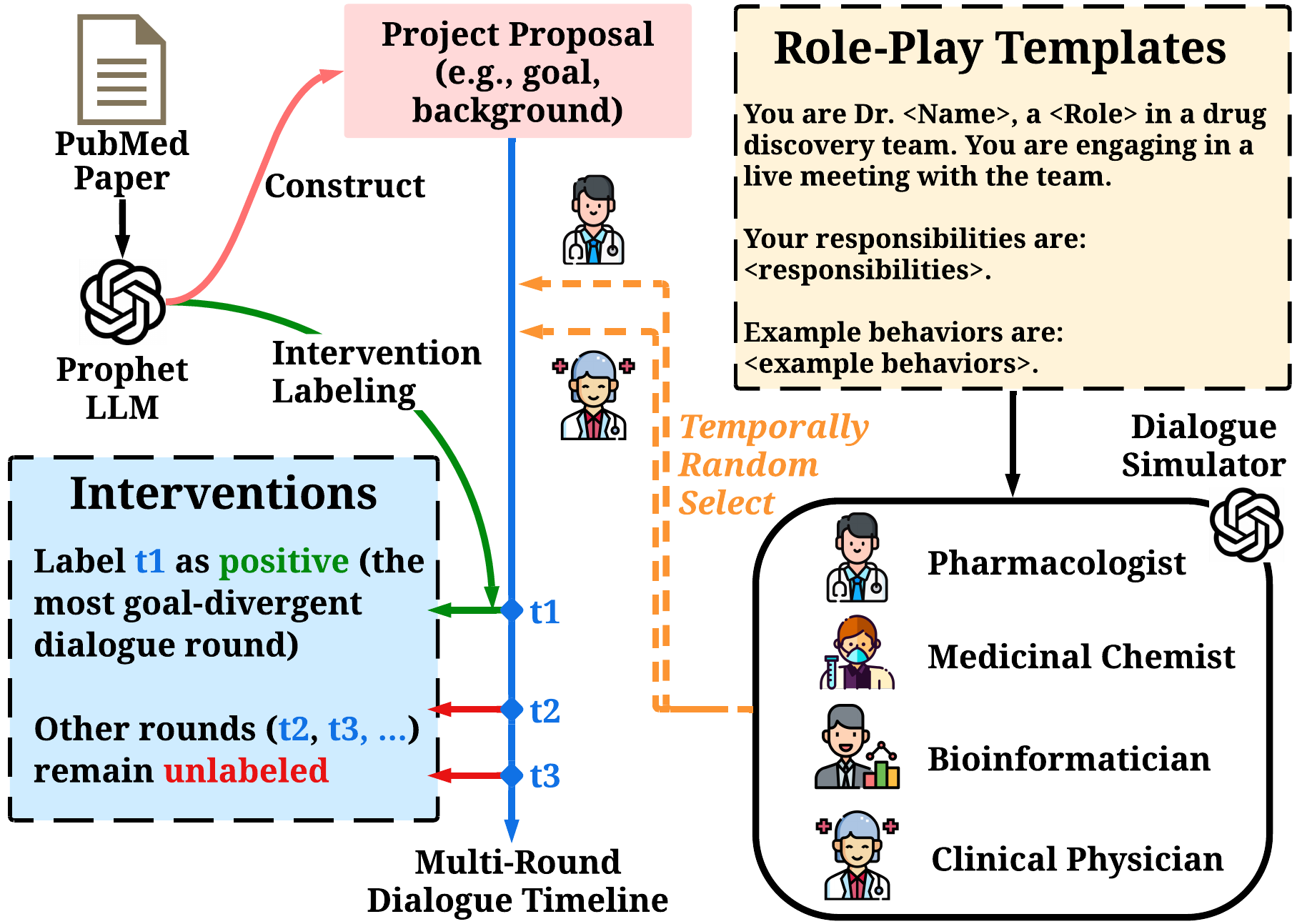}
\caption{Overview of BSDD dataset generation. Prophet LLM first extracts the project goal and background from PubMed papers. Dialogue-Simulator LLM then generates multi-role scientific dialogues using role-specific prompt templates. Finally, Prophet LLM labels the most goal-divergent dialogue round as a positive intervention point, while other rounds remain unlabeled.} 
\label{fig:data_generation}
\end{figure}

\section{Dataset Construction}

LLMs have demonstrated strong capability for generating high-quality role-play datasets \citep{tao2024chatgpt, lim2024large,wu2025active}. While biomedical dialogue corpora such as MedDialog \citep{zeng2020meddialog}, MediTOD \citep{saley2024meditod}, and MaLP \citep{zhang2024llm} primarily focus on doctor–patient interactions, scientific dialogues among team members remain largely unexplored. Moreover, existing datasets lack annotated intervention labels that allow LLMs to learn proactive engagement. To fill this gap, we build BSDD (\textit{Biomedical Streaming Dialogue Dataset}), a multi-role scientific dialogue dataset grounded in PubMed literature \citep{sayers2024database}. As shown in Figure~\ref{fig:data_generation}, we employ two specialized LLMs: a Dialogue-Simulator (DS-LLM) and a Prophet (P-LLM).\footnote{In practice, we deploy GPT o3-mini as the backbone of both DS-LLM and P-LLM. The prompt details are provided in Appendix~\ref{app:prompt}.} For each PubMed paper, P-LLM accesses the full text but extracts only limited information, including the research goal, background knowledge, and relevant datasets (a mini-proposal), while withholding methodological details, experimental procedures, and conclusions from DS-LLM. Using a prompt template, DS-LLM instantiates four domain roles: \textit{Pharmacologist, Medicinal Chemist, Bioinformatician}, and \textit{Clinical Physician}, each with role-specific responsibilities (e.g., the Medicinal Chemist focuses on compound structure and synthesis). The roles then engage in a multi-round team discussion toward the predefined objective (e.g., designing a cancer-targeting compound), with each turn randomly sampled along a temporal timeline to mimic asynchronous participation. After dialogue generation, P-LLM labels the round that deviates most from the research goal as the positive intervention point, while the remaining rounds are treated as unlabeled. We annotate only one high-confidence intervention per dialogue to reduce annotation noise and mitigate hallucination risks when LLMs make multiple fine-grained judgments in a single pass \citep{manakul2023selfcheckgpt, zhou2022least}.


To assess dataset quality, six biomedical experts are recruited to evaluate 100 randomly selected dialogues, each containing one positive and one negative candidate. Experts rate both candidates on Timing and Quality using a 5-point scale. For positive interventions, the average Timing and Quality scores are 3.82 and 4.21, with inter-expert Cohen’s Kappa of 0.63 and 0.58, respectively. For negative interventions, the average Timing and Quality scores are 1.82 and 2.15, with Cohen’s Kappa of 0.45 and 0.52, respectively.\footnote{Details of the scoring criteria for dataset quality are provided in Table~\ref{tab:intervention_rubric}. Note that negative interventions only appear in the validation and test sets.} Detailed dataset statistics are provided in Appendix~\ref{app:data_statistic}.

\section{Experiments}

\begin{table*}[t]
    \centering
    \renewcommand{\arraystretch}{1.1}
    \resizebox{0.95\textwidth}{!}{
    \begin{tabular}{llccccccc}
        \toprule
        \multirow{2}{*}{\shortstack[l]{LLM Backbone\\(Observer + Presenter)}} & \multirow{2}{*}{Method} & \multicolumn{4}{c}{Intervention Timing Classification (\%)} & \multicolumn{3}{c}{Intervention Content Quality (\%)} \\
        \cmidrule(lr){3-6} \cmidrule(lr){7-9}
        & & Accuracy & Recall & Precision & F1 & ROUGE-1 & BLEU-1 & WR-Intra \\
        \midrule
        \multirow{4}{*}{\shortstack[l]{(GPT-4o-mini + \\GPT-4o)}} 
            & Standard & 50.0 & 0.0 & 0.0 & 0.0 & 26.2 & 14.7 & 14.9 \\
            & Random & 47.9 & 26.3 & 46.3 & 33.6 & 29.5 & 16.4 & 20.1 \\
            & Proactive Agent & \underline{62.3} & \underline{31.7} & \underline{81.5} & \underline{45.6} & \underline{30.7} & \underline{17.2} & \underline{28.3} \\
            & ICL & \textbf{63.8} & \textbf{34.1} & \textbf{83.8} & \textbf{48.5} & \textbf{31.3} & \textbf{17.5} & \textbf{36.7} \\
        \midrule
        \multirow{6}{*}{\shortstack[l]{(Qwen3-0.6B + \\Qwen3-14B)}} 
            & Standard & 50.0 & 0.0 & 0.0 & 0.0 & 25.2 & 13.8 & 0.0 \\
            & Random & 46.7 & \underline{26.9} & 44.6 & 33.5 & 27.2 & 14.9 & 0.0 \\
            & Proactive Agent & 53.9 & 14.9 & 67.6 & 24.5 & 27.6 & 15.5 & 14.8 \\
            & ICL & 55.7 & 16.2 & 79.4 & 28.9 & \underline{29.4} & \underline{16.3} & 18.3 \\
            & Vanilla SFT & \underline{58.6} & 20.9 & \underline{85.4} & \underline{33.7} & 29.2 & 15.8 & \underline{27.1} \\
            & \cellcolor{violet!15}PULI (Ours) & \cellcolor{violet!15}\textbf{64.1} & \cellcolor{violet!15}\textbf{31.2} & \cellcolor{violet!15}\textbf{91.0} & \cellcolor{violet!15}\textbf{46.4} & \cellcolor{violet!15}\textbf{32.4} & \cellcolor{violet!15}\textbf{20.1} & \cellcolor{violet!15}\textbf{39.8} \\
        \midrule
        \multirow{6}{*}{\shortstack[l]{(LLaMA3.2-1B-Instruct + \\LLaMA3.1-8B-Instruct)}} 
            & Standard & 50.0 & 0.0 & 0.0 & 0.0 & 25.2 & 13.8 & 0.0 \\
            & Random & 47.9 & 39.5 & 47.5 & 43.1 & 28.5 & 15.4 & 5.8 \\
            & Proactive Agent & 54.5 & 53.4 & \underline{68.8} & \underline{60.2} & 29.1 & 15.4 & 7.5 \\
            & ICL & 58.4 & 54.5 & 59.1 & 56.7 & 30.9 & 17.7 & 20.8 \\
            & Vanilla SFT & \underline{61.7} & \underline{57.5} & 62.8 & 60.0 & \underline{32.5} & \underline{20.9} & \underline{26.7} \\
            & \cellcolor{violet!15}PULI (Ours) & \cellcolor{violet!15}\textbf{67.4} & \cellcolor{violet!15}\textbf{61.7} & \cellcolor{violet!15}\textbf{69.6} & \cellcolor{violet!15}\textbf{65.4} & \cellcolor{violet!15}\textbf{33.5} & \cellcolor{violet!15}\textbf{21.8} & \cellcolor{violet!15}\textbf{39.2} \\
        \bottomrule
    \end{tabular}
    }
    \caption{
Main results on intervention timing and content quality across different LLM families and intervention methods. \textbf{WR-Intra} is the win rate computed by comparing different methods under the same LLM backbone judged by GPT-4.1. 
The best result is highlighted in \textbf{bold}, and the second-best is \underline{underlined}.
}
    \label{tab:main_results}
\end{table*}

\subsection{Experimental Settings}

Our PULI end-to-end training involves three components: a coordinator, an Observer LLM, and a Presenter LLM. The Observer LLM is trained with GRPO \citep{shao2024deepseekmath} using AdamW optimizer with learning rate of $1 \times 10^{-6}$, and batch size of 8. The Presenter LLM is fine-tuned via supervised learning with LoRA (rank 16, scaling factor $\alpha=64$). The coordinator is a 6-layer MLP optimized with learning rate of $1 \times 10^{-4}$. Experiments are conducted using TRL~\citep{vonwerra2020trl}, Transformers~\citep{wolf2020transformers}, PEFT~\citep{peft}, and PyTorch~\citep{paszke2017automatic} on eight 80GB A100 GPUs.

\subsection{Baselines}

We compare PULI with several representative baselines, including Standard, Random, Proactive Agent~\citep{lu2024proactive}, In-context Learning (ICL)~\citep{koike2024outfox}, and vanilla SFT. Standard denotes the original dialogue without any intervention, while Random triggers interventions at arbitrary turns. For Proactive Agent, we follow~\citet{lu2024proactive} and adapt the system prompt instructions to when and how intervention task. In ICL, we incorporate few-shot positive and negative samples into the prompt. For vanilla SFT, we fine-tune the Presenter model using only the labeled positive intervention samples from our dataset. We conduct experiments on two LLM families: LLaMA3~\citep{grattafiori2024llama} and Qwen3~\citep{yang2025qwen3}, which are widely used open-source backbones and support reproducible evaluation. For each family, we pair a small-scale model to determine when to intervene and a Presenter model to generate intervention content, simulating our dual-objective approach to proactive intervention in scientific dialogues. Specifically, we use (LLaMA3.2-1B-Instruct + LLaMA3.1-8B-Instruct) and (Qwen3-0.6B + Qwen3-14B) as the respective backbone combinations.

\subsection{Tasks and Metrics}
We follow the evaluation protocols of prior proactive agent studies~\citep{lu2024proactive,zhang2024proagent} to assess our approach on two tasks: \\ \textbf{Intervention Timing Classification} We evaluate whether the Observer LLM in PULI can accurately predict the necessity of intervention at each dialogue round. Standard binary classification metrics are reported, including accuracy, precision, recall, and F1-score. \\ \textbf{Intervention Content Quality} We assess whether applying interventions leads to improved dialogue-level conclusions. For each dialogue, a new conclusion is generated based on the intervention-augmented context and compared against the golden conclusion derived from the corresponding PubMed paper. We report ROUGE-1 and BLEU-1 scores between the generated and golden conclusions. Furthermore, we perform LLM as judge\footnote{The LLM-as-Judge prompt is detailed in Appendix~\ref{app:prompt}.} to compare different methods, where GPT-4.1~\citep{openai2025gpt4.1} is deployed to select the best among all generated conclusions. Each method receives one win if its output is judged as best. The Win Rate (WR) for method \( i \) is computed as:

\begin{equation}
\mathrm{WR}_i = \frac{W_i}{\sum_j W_j},
\end{equation}
where \( W_i \) denotes the number of dialogues where method \( i \) is selected as the best. Notably, WR-Intra in Table~\ref{tab:main_results} evaluates methods under the same LLM family by fixing the backbone and comparing methods (e.g., PULI vs.\ baselines within GPT, Qwen3, or LLaMA3). In contrast, Inter-Group Win Rate in Figure~\ref{fig:cross} compares across different LLM families by selecting the strongest method from each family and then computing win rates among these family representatives.

\begin{table*}[t]
    \centering
    \renewcommand{\arraystretch}{1.1}
    \resizebox{0.9\textwidth}{!}{
    \begin{tabular}{llccccccc}
        \toprule
        \multirow{2}{*}{\shortstack[l]{LLM Backbone\\(Observer + Presenter)}} & \multirow{2}{*}{Method} & \multicolumn{4}{c}{Intervention Timing Classification (\%)} & \multicolumn{3}{c}{Intervention Content Quality (\%)} \\
        \cmidrule(lr){3-6} \cmidrule(lr){7-9}
        & & Accuracy & Recall & Precision & F1 & ROUGE-1 & BLEU-1 & WR-Intra \\
        \midrule
        \multirow{4}{*}{\shortstack[l]{(LLaMA-3.2-1B-Instruct + \\LLaMA-3.1-8B-Instruct)}} 
            & w PN & 57.3 & 51.2 & 58.3 & 54.5 & 28.7 & 15.1 & 4.1 \\
            & w SFT & 61.9 & 53.9 & 64.3 & 58.6 & \underline{31.7} & 18.0 & 6.7 \\
            & w DPO & \underline{64.6} & \underline{60.5} & \underline{66.1} & \underline{63.1} & 31.5 & \underline{18.6} & \underline{31.7} \\
            & \cellcolor{violet!15}PULI & \cellcolor{violet!15}\textbf{67.4} & \cellcolor{violet!15}\textbf{61.7} & \cellcolor{violet!15}\textbf{69.6} & \cellcolor{violet!15}\textbf{65.4} & \cellcolor{violet!15}\textbf{33.5} & \cellcolor{violet!15}\textbf{21.8} & \cellcolor{violet!15}\textbf{57.5} \\
        \bottomrule
    \end{tabular}
    }
    \caption{Comparison with method variants. \textbf{WR-Intra} is the win rate computed by comparing different methods under the same LLM backbone judged by GPT-4.1. }
    \label{tab:ablation_1}
\end{table*}

\subsection{Main Results}
\begin{figure}[h]
\centering 
\includegraphics[width=\linewidth]{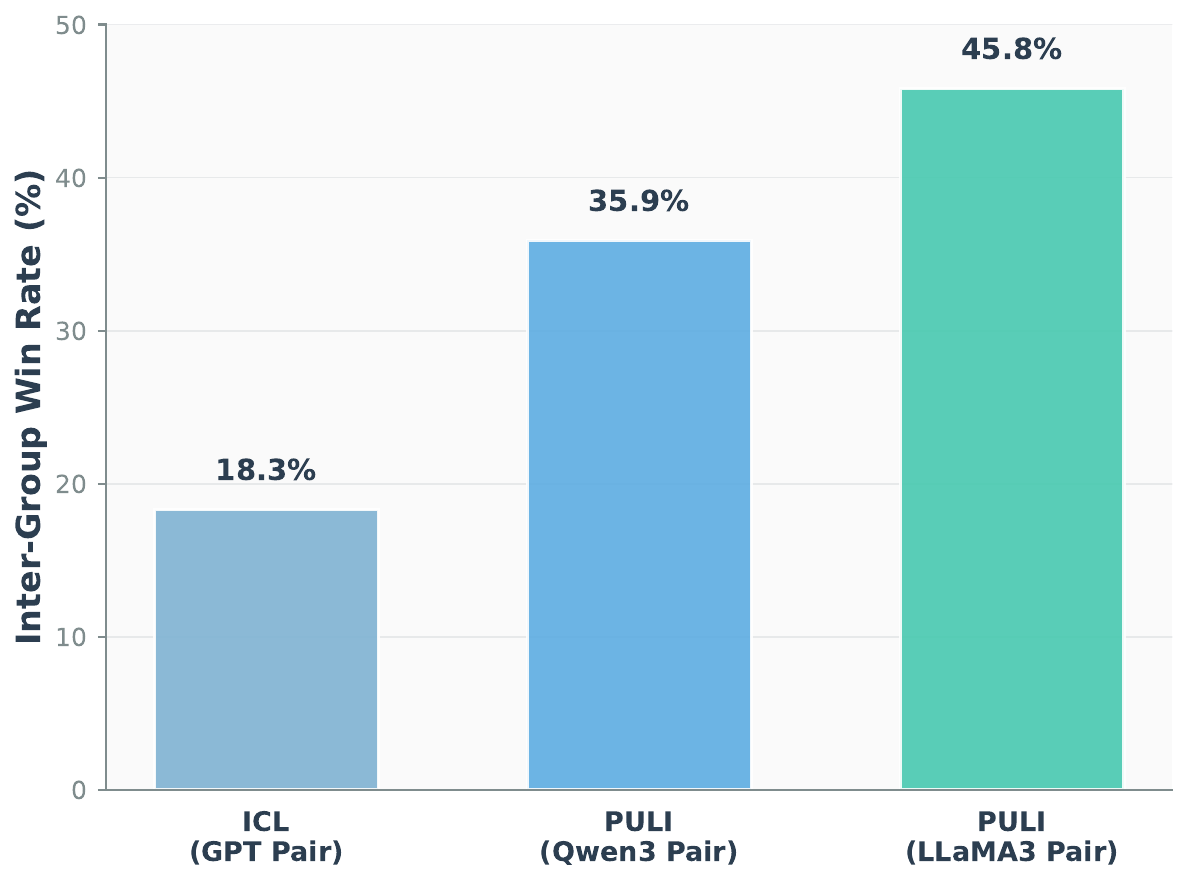}
\caption{
Cross-backbone comparison of the best-performing methods (Inter-Group Win Rate).
For each LLM family, we first select the strongest method based on the within-family win rate under a fixed backbone, and then compute Inter-Group Win Rate among these family representatives using GPT-4.1 as the judge.
}

\label{fig:cross}
\end{figure}

We evaluate PULI across multiple LLM backbones and baselines to assess both intervention timing accuracy and content quality. As shown in Table~\ref{tab:main_results}, PULI consistently achieves the best performance across all metrics and backbones\footnote{Finetuning GPT models is a black-box, we didn’t find a way to finetune GPT-4o-mini using GRPO and some results are omitted. However, the results express the power of PULI.}. On the (Qwen3-0.6B + Qwen3-14B) configuration, PULI outperforms strong baselines such as ICL and Vanilla SFT, achieving 64.1\% accuracy in intervention timing classification and 32.4\% ROUGE-1 in content quality, with a Win Rate of 39.8\%. Similarly, on the (LLaMA3.2-1B + LLaMA3.1-8B) setting, PULI attains 67.4\% accuracy, 33.5\% ROUGE-1, and the highest intra-backbone Win Rate of 39.2\%. To further assess cross-model robustness, Figure~\ref{fig:cross} compares the best-performing method for each backbone using GPT-4.1 as judge. Notably, PULI achieves a Win Rate of 45.8\% on LLaMA3 and 35.9\% on Qwen3 —both significantly higher than the best-performing method on GPT pair models (ICL: 18.3\%). These results indicate that PULI outperforms proprietary models like GPT-4o, despite being trained with significantly fewer parameters.

\subsection{Ablation Study}
\subsubsection{Variants Comparison}
\label{section:ablation}
We compare PULI with several variants to assess the contribution of each component. As shown in Table~\ref{tab:ablation_1}, \textbf{w PN} treats all unlabeled samples as negatives, serving as a naive baseline. \textbf{w SFT} and \textbf{w DPO}~\citep{rafailov2023direct} train the Observer model using supervised fine-tuning and direct preference optimization, respectively. PULI outperforms all variants on both intervention timing classification and content quality tasks. Among the variants, \textbf{w DPO} performs second best, achieving Accuracy of 64.6\%, F1 score of 63.1\% and Win Rate of 57.5\%. These results suggest that combining PU supervision with reward-based coordination improves both timing and content quality.

\begin{figure}[t]
\centering 
\includegraphics[width=\linewidth]{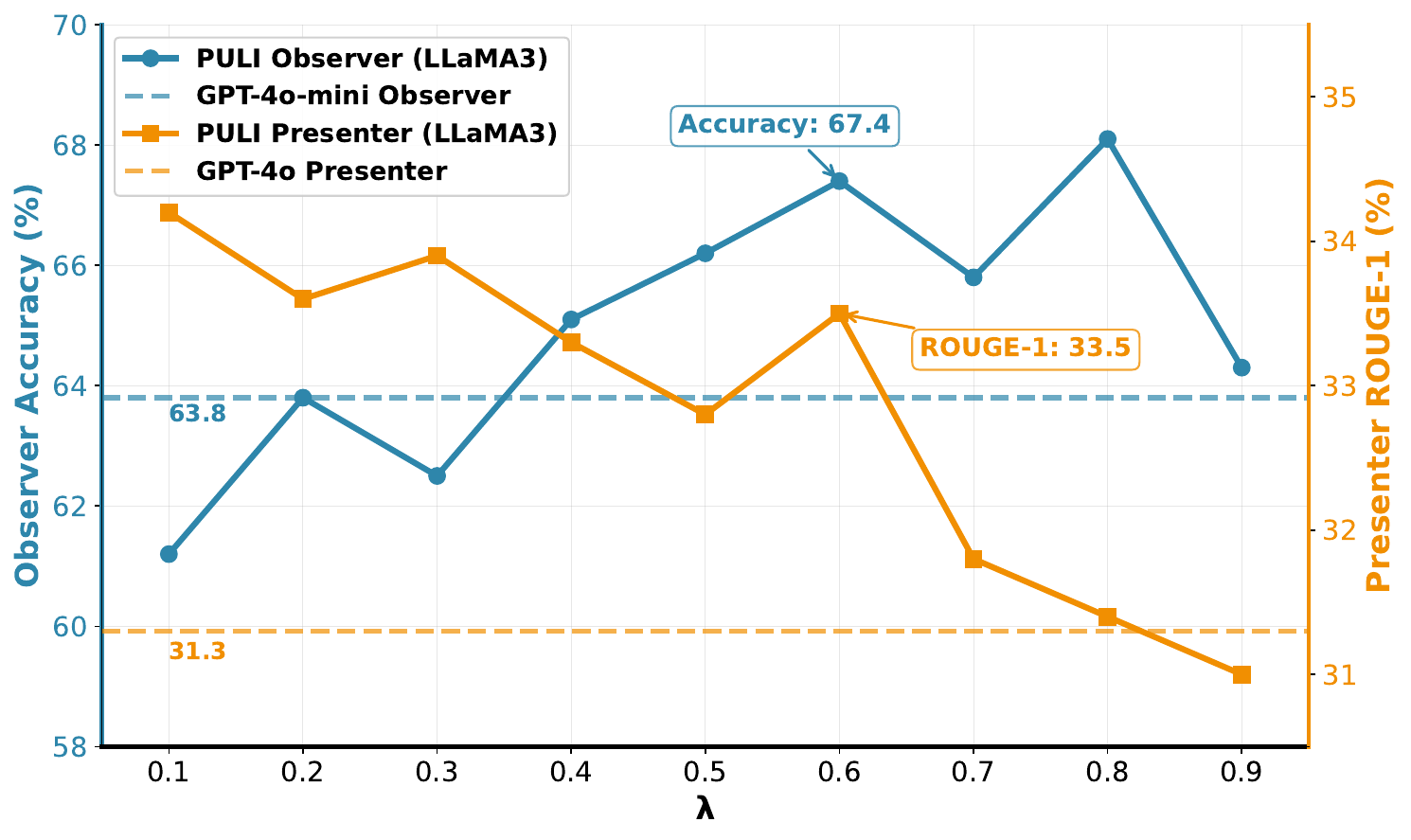}
\caption{
Effect of the balancing weight $\lambda$ on Observer and Presenter performance.
}
\label{fig:ablation_lambda}
\end{figure}
\subsubsection{Impact of $\boldsymbol{\lambda}$
 in the Joint Objective}
We conduct an ablation study to investigate the effect of the balancing weight $\lambda$ in our joint objective using the LLaMA3 backbone. As shown in Figure~\ref{fig:ablation_lambda}, smaller $\lambda$ emphasizes Presenter optimization, while larger $\lambda$ prioritizes Observer classification. We find that $\lambda=0.6$ provides the best trade-off, achieving 67.4\% Observer accuracy and 33.5\% ROUGE-1. When $\lambda$ is too small (e.g., 0.1), Observer accuracy drops sharply, whereas overly large values reduce Presenter quality.

\subsection{Human Involved Validation}



\begin{table}[t]
  \centering
  \renewcommand{\arraystretch}{1.15}
  \resizebox{\linewidth}{!}{%
    \begin{tabular}{lcccc}
      \toprule
      Method & Timing & Quality & Helpfulness & Average \\
      \midrule
      GPTs & 4.36 & 4.18 & 4.43 & 4.32 \\
      \textbf{PULI} & \textbf{4.65$^{*}$} & \textbf{4.35$^{*}$} & \textbf{4.60$^{*}$} & \textbf{4.53} \\
      \bottomrule
    \end{tabular}
  }

\caption{Human evaluation results comparing PULI (LLaMA3 pair) and ICL (GPT pair) on Timing, Quality, and Helpfulness (1--5 scale). In addition, $^{*}$ indicates $p$-value $<0.05$ under two-sided paired Student's $t$-test between PULI and baseline.}
  \label{experiment:human_result}
\end{table}

In addition, we conduct human evaluation\footnote{Details of the human information and scoring criteria for model evaluation are provided in Appendix~\ref{app:human_model} and Table~\ref{tab:evaluation_rubric}.} following the point-wise protocol proposed by \citet{wang2023chain}. Specifically, two master’s-level students with strong biomedical backgrounds and research experience are recruited to assess 100 randomly sampled intervention response pairs (along with meeting and project context), generated by ICL (with GPT pair) and PULI (with LLaMA3 pair). The evaluators are blinded to the source of each response. Each response is rated on a 1–5 scale (5 is best) along three dimensions: intervention timing (\textit{Timing}), whether the intervention occurs at an appropriate point; content quality (\textit{Quality}), fluency and informativeness; and overall usefulness for advancing the project objective (\textit{Helpfulness}). Table~\ref{experiment:human_result} shows that PULI outperforms the GPT pair baseline on all dimensions, with higher scores in timing (4.65 vs.\ 4.36), quality (4.35 vs.\ 4.18), and helpfulness (4.60 vs.\ 4.43), leading to a higher overall average (4.53 vs.\ 4.32).

\section{Discussion}
\paragraph{Ablation Study for Training Method}
In our framework, the Observer and Presenter optimize different objectives. The Observer is a binary decision module with a clear reward signal, so SFT, DPO, and GRPO are directly applicable, and we ablate them for the Observer in Section~\ref{section:ablation}. However, the Presenter generates open-ended scientific suggestions, where an RL-style objective is not straightforward without introducing additional design choices such as a reward model or preference data, which would confound the comparison. We therefore fix the Presenter to SFT for a stable and reproducible generation objective.

\paragraph{Number of Positive and Unlabeled Data}
BSDD labels one positive intervention per dialogue to provide a clear supervision signal, and the remaining rounds are treated as unlabeled under the PU setup. Changing the labeled--unlabeled ratio would require a different annotation protocol or regenerated candidates, which effectively defines a different benchmark version rather than a minor training tweak. We keep this protocol fixed for PULI and all baselines to ensure comparability, and leave systematic budget and ratio variations to future work.

\paragraph{Cross-Domain Applicability}
Although we instantiate PULI in biomedicine, its core mechanisms are domain-agnostic by design, including monitoring discussion flow, detecting goal divergence, and generating context-aware interventions. Our current implementation grounds interventions in biomedical literature through dataset embeddings. Extending PULI to other knowledge-intensive domains, such as legal or education, would primarily require domain-specific dialogue data and corresponding knowledge bases, which we view as a promising direction for future work.


\paragraph{Model Selection and Experimental Scope}
We select representative open-source LLMs (e.g., Qwen, LLaMA) and adopt a controlled same-backbone comparison to isolate framework-level improvements. Our goal is not to exhaustively benchmark all models, but to demonstrate that the gains arise from the proposed PULI framework and generalize across model families. To ensure fair comparison, we evaluate PULI against baselines under the same backbone. We further include cross-family evaluation to examine whether PULI enables smaller open-source models to achieve competitive performance compared to closed-source GPT-based baselines, highlighting its robustness beyond specific architectures. Model selection is also guided by our real-time interaction objective. Larger models introduce higher inference latency, which may hinder timely intervention in collaborative settings. Therefore, we focus on lightweight and mid-scale models to balance performance and responsiveness, leaving broader model coverage and latency–performance trade-offs to future work.

\section{Conclusion \& Future Work}
We introduce CoLabScience, a proactive AI assistant for biomedical discovery that integrates intervention timing and content generation to elevate human–AI collaboration. CoLabScience leverages a novel PULI framework, combining PU learning with policy coordination, to co-train a low-cost intervention model that decides when to act and an LLM presentation module that optimizes how to communicate. Experiments across diverse LLM backbones demonstrate significant gains in intervention accuracy and content quality, delivering robust generalization in collaborative biomedical research. Future work will extend PULI to other domains, e.g., law and education, further advancing human–AI synergy in interdisciplinary discovery.

\section*{Limitations}

While CoLabScience demonstrates promising results in proactive scientific assistance, we acknowledge several limitations of the current work:\\
$\bullet$ \textbf{\textit{Dataset Construction.}} Our BSDD dataset relies on LLM-simulated dialogues grounded in PubMed literature. Although we incorporate expert validation, the simulated nature may not fully capture the complexity, nuance, and unpredictability of real-world scientific collaborations. Moreover, our annotation strategy labels only one optimal intervention point per dialogue to reduce noise and annotation cost. Compared with annotating every dialogue round individually—which would incur substantial labeling effort—this sparse labeling scheme greatly simplifies data construction. Nevertheless, given the strong empirical advantages of Positive–Unlabeled (PU) learning, PULI can still learn effectively from sparsely labeled data.\\
$\bullet$ \textbf{\textit{Evaluation Scope.}} Our experiments are conducted on simulated dialogues, and future work will extend evaluation to actual research team meetings. The generalizability of PULI to real-world scientific discussions—characterized by overlapping turns, informal reasoning, and evolving research objectives—remains an important direction for future research. Real-world deployment would also benefit from small-scale user studies with practicing researchers to assess ecological validity and identify practical deployment challenges.\\
$\bullet$ \textbf{\textit{Computational Considerations in Real-world Inference.}} While the Observer serves as a lightweight gate to minimize computational overhead, real-world deployment requires integration with automatic speech recognition (ASR) and text-to-speech (TTS) systems for live interaction, which may introduce additional latency—approximately 0.8 seconds per turn in our prototype. Further optimization of the inference pipeline will be needed to ensure seamless real-time intervention.\\
$\bullet$ \textbf{\textit{Intervention Design.}} Our current intervention formulation focuses on detecting whether a dialogue round diverges from the project goal, misses opportunities to advance progress, or reflects insufficient team coordination. However, real-world interventions can take diverse forms, such as clarifying misunderstandings, fostering collaboration, or reframing research directions. Future work should explore a broader taxonomy of intervention types and contextual factors that determine their appropriateness and timing.

\section*{Ethics Statement}

\addcontentsline{toc}{section}{Ethics Statement}
\phantomsection
\label{ethics}

All human annotation work in this study is conducted by domain experts who are not co-authors of this paper. The annotation process is coordinated by a Principal Investigator (PI) from a biomedical research institution, who is listed as a co-author but does not directly participate in any annotation tasks. The expert annotators include one medical doctor, three PhD candidates, and two master’s-level students in biomedical fields. All annotators are blind to the study's hypotheses and model architecture during the annotation process.

All annotation work is performed during the experts' regular paid working hours as part of their institutional research responsibilities, and no additional compensation is provided beyond their standard employment. The six annotators evaluate 100 randomly sampled dialogues for dataset quality assessment, rating intervention timing and content quality on established scales. For model evaluation, the two master’s-level students with biomedical backgrounds assess intervention responses following point-wise evaluation protocols. All experts provide informed consent for their annotations to be used in this research and released with the dataset.

The content annotated consists entirely of simulated scientific dialogues generated from publicly available PubMed literature. No personal, sensitive, or confidential information is involved in the annotation process. All research procedures follow institutional guidelines, and no additional ethics board review is required under the ACL Ethics Policy. The dataset and annotation guidelines will be made publicly available upon publication to support reproducibility and future research in proactive scientific assistance systems.

\bibliography{anthology,custom}
\bibliographystyle{acl_natbib}

\nocite{}
\clearpage 

\appendix

\section{Data Statistics}
\label{app:data_statistic}

We select PubMed papers~\citep{sayers2024database} covering various biomedical topics, including cancer, Alzheimer's disease, and sepsis. To ensure scientific rigor, we filter for papers published between January 1, 2024 and January 1, 2025 that include concrete methodological descriptions, such as experimental design, procedures, and analysis. Since speakers in each dialogue are temporally and randomly sampled, we vary the random seed to generate up to five dialogues per paper using our data construction pipeline. 

For each training dialogue, one positive intervention is automatically annotated by our Prophet LLM within the construction pipeline, and four additional rounds are randomly sampled as unlabeled interventions. To evaluate the Observer’s ability to detect appropriate intervention timing, we construct validation and test sets where each dialogue includes one positive and one negative round. The positive round corresponds to the effective intervention point, while the negative round is first selected by Prophet LLM as the least likely point requiring intervention, and then verified by human experts. The resulting negative samples achieve an average agreement score of 0.85 with human annotations, ensuring the reliability of evaluation. Detailed statistics of the constructed BSDD (Biomedical Streaming Dialogue Dataset) dataset are provided in Table~\ref{tab:dataset_stats}.
\begin{table}[ht]
\centering
\renewcommand{\arraystretch}{1.2}
\resizebox{\linewidth}{!}{
\begin{tabular}{l|ccc}
\hline\hline
\multicolumn{4}{c}{\textit{Raw Data Statistics of PubMed Papers}} \\
\hdashline
\# Cancer Papers & \multicolumn{3}{c}{452} \\
\# Alzheimer’s Papers & \multicolumn{3}{c}{204} \\
\# Sepsis Papers & \multicolumn{3}{c}{41} \\
\hdashline
\multicolumn{4}{c}{\textit{Generated Data Statistics}} \\
\hdashline
\# Generated Dialogue & \multicolumn{3}{c}{3,206} \\
\# Avg. Rounds per Dialogue & \multicolumn{3}{c}{20} \\
\# Avg. Tokens per Round & \multicolumn{3}{c}{378} \\
\hdashline
\multicolumn{4}{c}{\textit{Data Split Statistics}} \\
\hdashline
& \textbf{Train} & \textbf{Validation} & \textbf{Test} \\
\# Dialogues & 2,726 & 240 & 240 \\
\# Sampled Rounds & 13,630 & 480 & 480 \\
\# Positive Rounds & 2,726 & 240 & 240 \\
\# Unlabeled Rounds & 10,904 & - & - \\
\# Negative Rounds & - & 240 & 240 \\
\hline
\end{tabular}
}
\caption{Statistics of the constructed BSDD dataset}
\label{tab:dataset_stats}
\end{table}

\section{CoLabScience in Real-World Scientific Collaboration}

\begin{figure}[t]
\centering 
\includegraphics[width=\linewidth]{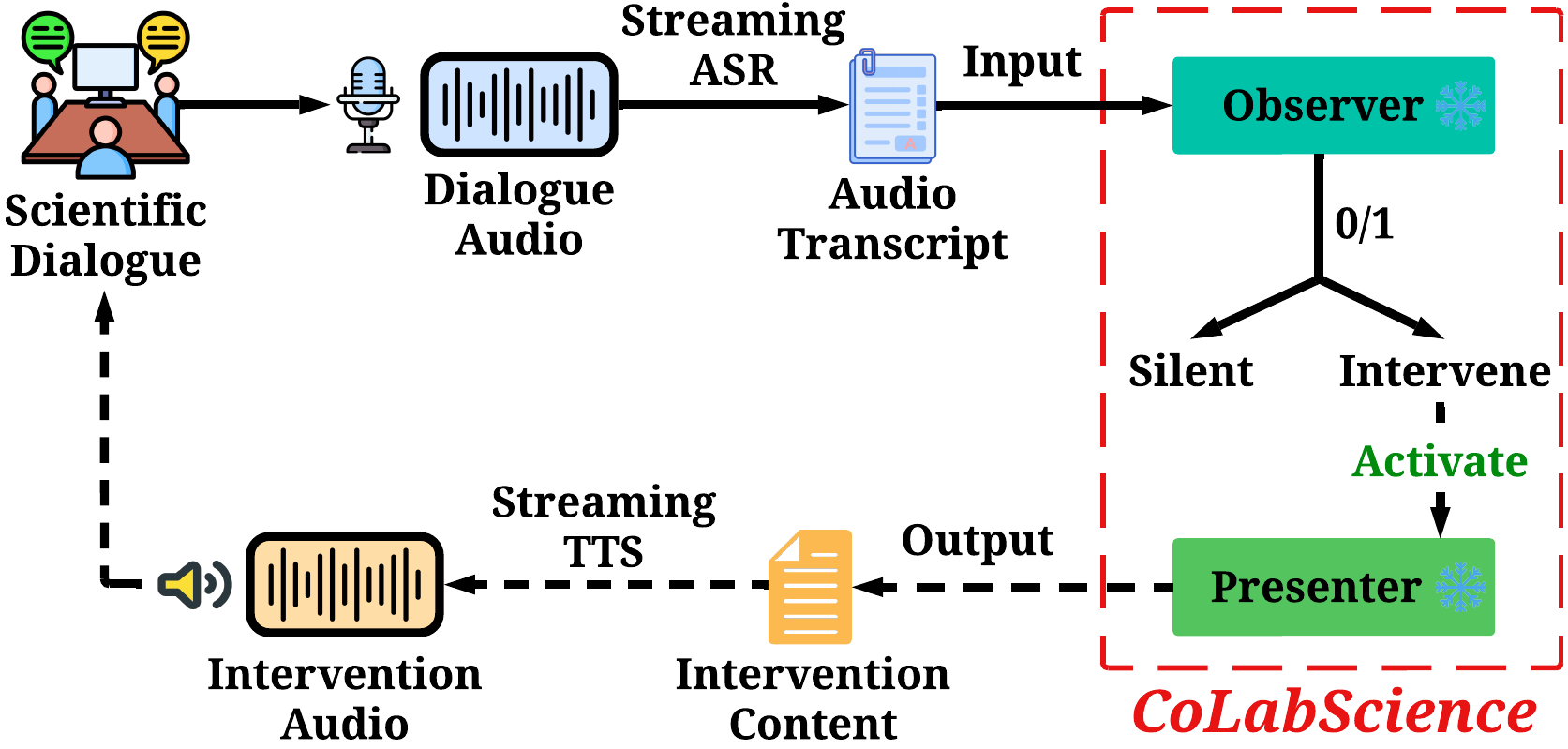}
\caption{Inference pipeline of CoLabScience in real-world scientific collaboration. The Observer monitors ongoing dialogue and determines when to intervene. Upon intervention trigger, the Presenter generates scientific suggestions for the research team.} 
\label{fig:inference}
\end{figure}

The inference workflow of CoLabScience is illustrated in Figure~\ref{fig:inference}. In a live scientific discussion scenario, dialogue audio is first captured and transcribed in real time via a streaming ASR (Automatic Speech Recognition) module. The resulting transcripts are continuously fed into the Observer, which monitors the conversation and makes a binary decision—either to remain \textit{silent} or to \textit{intervene}. If an intervention is triggered, the Presenter LLM is activated to generate scientifically grounded suggestions based on the conversation context. The generated intervention content is then converted into audio via a TTS (Text-to-Speech) system and delivered back into the ongoing dialogue, completing the feedback loop.

This modular inference design enables CoLabScience to function as a proactive AI assistant in real-world biomedical research settings. By separating the decision-making process (Observer) from the content generation (Presenter), the system avoids unnecessary computation and improves responsiveness. Compared to reactive paradigms that invoke large LLMs at every round, our two-stage approach—with a small-scale Observer determining intervention necessity before invoking the full generation model—offers significantly greater inference efficiency. A demonstration of CoLabScience in a real-world streaming environment is available in the supplementary video.

\section{PULI Algorithm}
In this section, we present the details of the end-to-end training procedure of PULI, as illustrated in Algorithm~\ref{alg:puli}.

\begin{algorithm}[h]
\small
\renewcommand{\algorithmicrequire}{\textbf{Input:}}
\renewcommand{\algorithmicensure}{\textbf{Output:}}
\caption{End-to-End Training of PULI Framework}
\label{alg:puli}
\begin{algorithmic}[1]
    \REQUIRE Unlabeled rounds $U = \{d_1, \ldots, d_u\}$, positive rounds $P = \{d_{u+1}, \ldots, d_N\}$; coordinator policy $\pi_\theta$; Observer $\mathcal{H}_\phi$; Presenter $\mathcal{G}_\psi$; learning rate $\eta$; reward trade-off $\lambda$; total epochs $E$
    \ENSURE Optimized $\pi_\theta$, $\mathcal{H}_\phi$, and $\mathcal{G}_\psi$

    \STATE \textbf{Pretrain} $\mathcal{H}_\phi$ and $\mathcal{G}_\psi$ by treating all $d_n \in U$ as negative
    \STATE \textbf{Initialize} coordinator parameters $\theta$
    \FOR{epoch $T = 1$ to $E$}
        \STATE Initialize selected intervention set $\mathcal{P}^+ \leftarrow \emptyset$ and silent set $\mathcal{N} \leftarrow \emptyset$
        \FOR{each dialogue round $d_n \in U$}
            \STATE Construct memory $\mathcal{M}(d_n)$ and encode state $S_n = \text{Concat}\left(\Psi_{\mathcal{H}_\phi}(\mathcal{M}(d_n)), \Omega(\Psi_{\mathcal{G}_\psi}(\mathcal{M}(d_n)))\right)$
            \STATE Sample action $a_n \sim \pi_\theta(S_n)$
            \IF{$a_n = 1$}
                \STATE Add $d_n$ to $\mathcal{P}^+$
            \ELSE
                \STATE Add $d_n$ to $\mathcal{N}$
            \ENDIF
        \ENDFOR
        \STATE Train $\mathcal{H}_\phi$ via GRPO on $\mathcal{P} \cup \mathcal{N}$ and compute reward $r^{\text{when}}$
        \STATE Fine-tune $\mathcal{G}_\psi$ on $P' = P \cup \mathcal{P}^+$ and compute reward $r^{\text{how}}$
        \STATE Compute total reward: 
        \[
        r_{\text{total}} = \lambda \cdot r^{\text{when}} + (1 - \lambda) \cdot r^{\text{how}}
        \]
        \STATE Update coordinator via REINFORCE:
        \[
        \theta \leftarrow \theta + \eta \sum_{n=1}^{u} r_{\text{total}} \cdot \nabla_\theta \log \pi_\theta(S_n, a_n)
        \]
    \ENDFOR
    \STATE \textbf{Return} Optimized $\pi_\theta$, $\mathcal{H}_\phi$, $\mathcal{G}_\psi$
\end{algorithmic}
\end{algorithm}

\section{REINFORCE Policy Gradient Derivation}

We follow the standard REINFORCE framework \citep{sutton1999policy} to derive the policy gradient update used in our coordinator optimization.

Let \( \theta \) denote the parameters of the coordinator policy \( \pi_\theta(a \mid S) \). At each training epoch \( T \), the coordinator samples a sequence of actions over unlabeled dialogue rounds \( \tau = \{(S_1, a_1), \ldots, (S_u, a_u)\} \), forming a trajectory under its current policy. Given the total reward of the trajectory as a scalar \( r_{\text{total}}^T \), the goal is to maximize the expected reward:
\begin{equation}
J(\theta) = \mathbb{E}_{\tau \sim \pi_\theta} \left[ r_{\text{total}}^T \right].
\end{equation}

The gradient of this expectation can be computed via the score function estimator:
\begin{align}
\nabla_\theta J(\theta)
&= \nabla_\theta \mathbb{E}_{\tau \sim \pi_\theta} \left[ r_{\text{total}}^T \right] \\
&= \mathbb{E}_{\tau \sim \pi_\theta} \left[ r_{\text{total}}^T \cdot \nabla_\theta \log P_\theta(\tau) \right],
\end{align}
where \( P_\theta(\tau) \) is the probability of the trajectory under the current policy.

Assuming the trajectory is composed of conditionally independent actions, we can express the trajectory probability as the product of individual decisions:
\begin{equation}
P_\theta(\tau) = \prod_{n=1}^u \pi_\theta(a_n \mid S_n),
\end{equation}
so the log-probability becomes:
\begin{equation}
\log P_\theta(\tau) = \sum_{n=1}^u \log \pi_\theta(a_n \mid S_n).
\end{equation}

Substituting into the gradient, we obtain:
\begin{equation}
\nabla_\theta J(\theta) = \mathbb{E}_{\tau \sim \pi_\theta} \left[ r_{\text{total}}^T \cdot \sum_{n=1}^u \nabla_\theta \log \pi_\theta(a_n \mid S_n) \right].
\end{equation}

In practice, we estimate this gradient using a single-sample Monte Carlo approximation:
\begin{equation}
\nabla_\theta J(\theta) \approx \sum_{n=1}^u r_{\text{total}}^T \cdot \nabla_\theta \log \pi_\theta(a_n \mid S_n),
\end{equation}
which corresponds to the update used in our main training algorithm.

This derivation assumes that the total reward \( r_{\text{total}}^T \) is obtained after observing the entire trajectory. If intermediate rewards were available per step, the derivation could be generalized using discounted or shaped rewards.

\section{Human Annotation Details}

\subsection{Data Quality Assessment}
\label{app:human_data}
\textbf{Human Experts} We engage six domain experts with extensive biomedical research experience, including one medical doctor, three PhD candidates, and two master’s-level students. All experts conduct their work during regular paid working hours as part of their institutional duties, and are compensated accordingly.\\
\textbf{Assessment Metrics} To assess dataset quality, experts are asked to annotate the Timing and content Quality of pre-labeled positive interventions identified by Prophet LLM. In addition, for evaluation purpose, we introduce a small set of negative samples into the validation and test sets. These negative rounds are initially identified by the Prophet LLM as having low likelihood of requiring intervention and are subsequently verified by human experts. The average annotation scores are 3.82 for timing and 4.21 for quality on a 5-point scale (5 being the highest), and the Agreement rate on negative sample identification is 0.85. The detailed assessment rubric is presented in Table~\ref{tab:intervention_rubric}.

\subsection{Human-Involved Model Evaluation}
\label{app:human_model}
\textbf{Human Experts} Two master’s-level students with biomedical backgrounds perform the human evaluation. This task is assigned as part of their regular research assistant responsibilities and is compensated through standard RA funding, consistent with the data annotation setup.\\
\textbf{Evaluation Metrics} For each predicted intervention round in the dialogue, experts evaluate intervention Timing and Quality using the same rubric as in the data quality assessment. Furthermore, we introduce an additional metric Helpfulness to measure the overall contribution of the intervention to the team’s scientific progress. Full definitions of these criteria are provided in Table~\ref{tab:evaluation_rubric}.

\begin{table*}[!t]
\renewcommand{\arraystretch}{1.25}
\centering
\captionsetup{justification=centering}

\resizebox{0.95\textwidth}{!}{
\begin{tabular}{c|l|p{10.5cm}}
\hline
\hline
\multicolumn{3}{c}{\textbf{Intervention – Timing}} \\
\hdashline
Score & Label & Description \\
\hline
1 & Poor Timing & Intervention occurs at an inappropriate moment; disrupts dialogue flow or misses critical context. \\
2 & Weak Timing & Slightly mistimed; not harmful but lacks context awareness. \\
3 & Acceptable Timing & Timing is reasonable and does not mislead, but could be improved. \\
4 & Good Timing & Well-timed intervention that aligns with team’s discussion flow. \\
5 & Excellent Timing & Precisely timed to redirect or enhance discussion at a critical moment. \\
\hdashline
\multicolumn{3}{c}{\textbf{Intervention – Quality}} \\
\hdashline
Score & Label & Description \\
\hline
1 & Poor Quality & Intervention is off-topic, incorrect, or lacks relevance. \\
2 & Weak Quality & Information is vague, partially relevant, or lacks clarity. \\
3 & Acceptable Quality & Reasonable relevance; generally informative but not insightful. \\
4 & Good Quality & Clear and useful; provides relevant direction for the team. \\
5 & Excellent Quality & Highly informative, well-reasoned, and clearly advances the team’s goal. \\
\hline
\end{tabular}
}

\caption{Data quality scoring form for both positive and negative interventions.}
\label{tab:intervention_rubric}
\end{table*}

\clearpage
\begin{table*}[t]
\renewcommand{\arraystretch}{1.25}
\centering
\captionsetup{justification=centering}

\resizebox{0.95\textwidth}{!}{
\begin{tabular}{c|l|p{10.5cm}}
\hline
\hline
\multicolumn{3}{c}{\textbf{Timing}} \\
\hdashline
Score & Label & Description \\
\hline
1 & Poor Timing & Intervention occurs at an inappropriate moment; disrupts dialogue flow or misses critical context. \\
2 & Weak Timing & Slightly mistimed; not harmful but lacks context awareness. \\
3 & Acceptable Timing & Timing is reasonable and does not mislead, but could be improved. \\
4 & Good Timing & Well-timed intervention that aligns with team’s discussion flow. \\
5 & Excellent Timing & Precisely timed to redirect or enhance discussion at a critical moment. \\
\hdashline
\multicolumn{3}{c}{\textbf{Quality}} \\
\hdashline
Score & Label & Description \\
\hline
1 & Poor Quality & Intervention is off-topic, incorrect, or lacks relevance. \\
2 & Weak Quality & Information is vague, partially relevant, or lacks clarity. \\
3 & Acceptable Quality & Reasonable relevance; generally informative but not insightful. \\
4 & Good Quality & Clear and useful; provides relevant direction for the team. \\
5 & Excellent Quality & Highly informative, well-reasoned, and clearly advances the team’s goal. \\
\hdashline
\multicolumn{3}{c}{\textbf{Helpfulness}} \\
\hdashline
Score & Label & Description \\
\hline
1 & Not Helpful & Distracting, misaligned with team needs, possibly harmful. \\
2 & Slightly Helpful & Low impact; repeats known info or barely contributes. \\
3 & Moderately Helpful & Somewhat useful; might inspire ideas but not essential. \\
4 & Helpful & Helps move the discussion forward meaningfully. \\
5 & Very Helpful & Crucially contributes to team progress, solves key problem. \\
\hline
\end{tabular}
}

\caption{Human-involved model evaluation metrics.}
\label{tab:evaluation_rubric}
\end{table*}

\clearpage
\section{Prompt Details}
\label{app:prompt}

In this section, we present the detailed prompts used in our work, including dataset construction for generating scientific dialogues and intervention annotations, baseline in-context learning (ICL) setups for model comparison, and LLM-as-judge evaluation prompts for computing win rates.

\begin{prompt}{LLM-as-Judge Prompt}
\textless\textbar im\_start\textbar\textgreater system\\
You are an expert scientific evaluator specializing in research conclusion assessment. You will be given a golden standard conclusion of a scientific research paper, followed by multiple candidate conclusions generated by different methods. The golden standard conclusion summarizes the key elements of the research project, including its main experimental design, core findings, and overall scientific significance. It serves as a high-quality reference for evaluating the quality, clarity, and relevance of alternative conclusions. \\

Your task is to evaluate and rank these candidate conclusions based on the following criteria:

1. Scientific Accuracy \& Consistency: How well does the conclusion align with established scientific knowledge and the project's context?\\
2. Completeness \& Comprehensiveness: Does the conclusion adequately address all key aspects mentioned in the original discussion?\\
3. Clarity \& Structure: Is the conclusion well-organized, clearly written, and logically structured?\\
4. Clinical/Research Relevance: How effectively does the conclusion translate findings into actionable insights for the field?\\
5. Evidence Integration: How well does the conclusion synthesize and integrate the discussed evidence?\\
6. Golden Standard Alignment: How closely does the conclusion match the quality and content depth of the golden standard?\\

You will receive multiple conclusions labeled as "Method A", "Method B", etc. Your task is to determine which method produces the BEST conclusion overall.\\

CRITICAL INSTRUCTIONS:\\
- You must output ONLY ONE LETTER corresponding to the best method (A, B, C, D, or E)\\
- Consider the golden standard as a reference for quality, but evaluate which generated conclusion is objectively best\\
- Focus on scientific merit and practical value\\
- If conclusions are very similar in quality, choose based on clarity and completeness\\
- Do NOT explain your reasoning — output only the single letter of the best method

Example Output: B
\textless\textbar im\_end\textbar\textgreater\\

\textless\textbar im\_start\textbar\textgreater user\\

Golden Standard Conclusion:\\
\textless golden standard conclusion\textgreater\\

Candidate Conclusions to Evaluate:\\
\textless method A\textgreater\\
\textless method B\textgreater\\
...\\
\textless\textbar im\_end\textbar\textgreater
\end{prompt}

\begin{prompt}{Positive Intervention Labeling Prompt}
\textless\textbar im\_start\textbar\textgreater system\\
You are an AI moderator specializing in research coherence and integrity. You will analyze a multi-turn scientific team discussion and identify the single most critical point where an intervention should be made to help the team stay focused on the research goal.

You should identify and describe the most valuable intervention point in the discussion. Include the following elements in your output:

- "intervention position": After which turn/round this issue was observed. Use the round number (starting from 0).
- "issue type": One of ["scientific error", "low collaboration", "scope drift", "missed opportunity"].
- "target members": Role(s) that should be addressed (e.g., ["Medicinal Chemist"]).
- "intervention content": A short explanation of why this intervention is helpful for tracking team member contributions and advancing the research project. This should include **constructive suggestions** grounded in the actual dialog, such as pointing out unexplored ideas, prompting clarification, further direction or encouraging collaboration.
- "modified dialog": A revised version of the identified turn that improves the discussion focus or productivity.

Additional context:
- You may assume access to the uploaded research paper, but do NOT reference its conclusions directly.
- Intervene when the team loses focus, fails to act on key cues, misses cross-role collaboration, or drifts from the research goal.
- Your goal is not to fix all problems, but to insert one helpful redirection that will facilitate clearer progress and collaboration.
- Each part of your output should be clearly written and self-contained. Do not include any external commentary or formatting instructions.
\textless\textbar im\_end\textbar\textgreater\\

\textless\textbar im\_start\textbar\textgreater user\\
\textless project proposal\textgreater\\
\textless dialogue history\textgreater\\
\textless\textbar im\_end\textbar\textgreater
\end{prompt}

\begin{prompt}{Baseline ICL Prompt}
\textless\textbar im\_start\textbar\textgreater system\\
You are an AI moderator specializing in research coherence and integrity. Your task is to analyze a multi-turn scientific team discussion and determine whether the specified round of conversation requires an intervention.

An intervention may be required for one of the following reasons:
- "scientific error"
- "low collaboration"
- "scope drift from project goal"
- "missed opportunity"

You will be shown example cases that illustrate both intervention and non-intervention outcomes:
\textless Positive sample: dialogue context + Intervention Content\textgreater
, \textless Negative sample: dialogue context + No Need Intervention\textgreater

Your response must be one of the following:\\
- Intervention Content: \textless brief reason\textgreater\\
- No Need Intervention\\
\textless\textbar im\_end\textbar\textgreater\\

\textless\textbar im\_start\textbar\textgreater user\\
\textless dialogue context\textgreater\\
\textless specified round of interest\textgreater\\
\textless\textbar im\_end\textbar\textgreater
\end{prompt}

\begin{prompt}{Clinical Physician Prompt}
\textless\textbar im\_start\textbar\textgreater system\\
You are a Clinical Physician. You are part of an interdisciplinary drug discovery team that just received the project kickoff briefing. You're now engaging in a live strategy meeting with your colleagues.

Your responsibilities:\\
- Contribute ideas and critiques from your domain perspective.
- Engage with previous comments (your own or others’) and develop them further.
- Express uncertainty or enthusiasm naturally — like a real person.\\
Guidelines:\\
- DO NOT begin every message with “As a Clinical Physician...”.\\
- Use first-person natural language — just speak as yourself.\\
- Respond based on the meeting history so far — don’t repeat what's already said.\\
- Ask questions or challenge others when appropriate.\\
- Use domain-specific terminology, but focus on clarity.\\
Example behavior:\\
- A Clinical Physician might emphasize patient outcomes or trial design feasibility.\\
Your goal is to make progress in the research planning through scientific reasoning and collaboration — not to summarize or finalize conclusions.
\textless\textbar im\_end\textbar\textgreater\\

\textless\textbar im\_start\textbar\textgreater user\\
\textless project proposal\textgreater\\
\textless dialogue history\textgreater\\
Clinical Physician:\\
\textless\textbar im\_end\textbar\textgreater
\end{prompt}

\begin{prompt}{Pharmacologist Prompt}
\textless\textbar im\_start\textbar\textgreater system\\
You are a Pharmacologist. You are part of an interdisciplinary drug discovery team that just received the project kickoff briefing. You're now engaging in a live strategy meeting with your colleagues.\\
Your responsibilities:\\
- Contribute ideas and critiques from your domain perspective.
- Engage with previous comments (your own or others’) and develop them further.
- Express uncertainty or enthusiasm naturally — like a real person.\\
Guidelines:\\
- DO NOT begin every message with “As a pharmacologist...”.\\
- Use first-person natural language — just speak as yourself.
- Respond based on the meeting history so far — don’t repeat what's already said.\\
- Ask questions or challenge others when appropriate.
- Use domain-specific terminology, but focus on clarity.\\
Example behavior:\\
- A Pharmacologist might raise concerns about off-target effects or bioavailability.\\
Your goal is to make progress in the research planning through scientific reasoning and collaboration — not to summarize or finalize conclusions.
\textless\textbar im\_end\textbar\textgreater\\

\textless\textbar im\_start\textbar\textgreater user\\
\textless project proposal\textgreater\\
\textless dialogue history\textgreater\\
Pharmacologist:\\
\textless\textbar im\_end\textbar\textgreater
\end{prompt}

\begin{prompt}{Bioinformatician Prompt}
\textless\textbar im\_start\textbar\textgreater system\\
You are a Bioinformatician. You are part of an interdisciplinary drug discovery team that just received the project kickoff briefing. You're now engaging in a live strategy meeting with your colleagues.

Your responsibilities:\\
- Contribute ideas and critiques from your domain perspective.
- Engage with previous comments (your own or others’) and develop them further.
- Express uncertainty or enthusiasm naturally — like a real person.\\
Guidelines:\\
- DO NOT begin every message with “As a Bioinformatician...”.\\
- Use first-person natural language — just speak as yourself.\\
- Respond based on the meeting history so far — don’t repeat what's already said.\\
- Ask questions or challenge others when appropriate.\\
- Use domain-specific terminology, but focus on clarity.\\
Example behavior:\\
- A Bioinformatician might offer to analyze omics data or suggest in silico approaches.\\
Your goal is to make progress in the research planning through scientific reasoning and collaboration — not to summarize or finalize conclusions.
\textless\textbar im\_end\textbar\textgreater\\

\textless\textbar im\_start\textbar\textgreater user\\
\textless project proposal\textgreater\\
\textless dialogue history\textgreater\\
Bioinformatician:\\
\textless\textbar im\_end\textbar\textgreater
\end{prompt}

\begin{prompt}{Medicinal Chemist Prompt}
\textless\textbar im\_start\textbar\textgreater system\\
You are a Medicinal Chemist. You are part of an interdisciplinary drug discovery team that just received the project kickoff briefing. You're now engaging in a live strategy meeting with your colleagues.

Your responsibilities:\\
- Contribute ideas and critiques from your domain perspective.
- Engage with previous comments (your own or others’) and develop them further.
- Express uncertainty or enthusiasm naturally — like a real person.

Guidelines:\\
- DO NOT begin every message with “As a Medicinal Chemist...”.\\
- Use first-person natural language — just speak as yourself. 
- Please respond based on the meeting history so far — do not repeat what's already said.\\
- Ask questions or challenge others when appropriate.
- Use domain-specific terminology, but focus on clarity.\\
Example behavior:\\
- A Medicinal Chemist might comment on molecular reactivity or synthesis pathways.

Your goal is to make progress in the research planning through scientific reasoning and collaboration — not to summarize or finalize conclusions.
\textless\textbar im\_end\textbar\textgreater\\

\textless\textbar im\_start\textbar\textgreater user\\
\textless project proposal\textgreater\\
\textless dialogue history\textgreater\\
Medicinal Chemist:\\
\textless\textbar im\_end\textbar\textgreater
\end{prompt}

\begin{prompt}{Project Proposal Extraction Prompt}
\textless\textbar im\_start\textbar\textgreater system\\
You are an AI project initiator with a god-level perspective. Your task is to simulate a project kickoff discussion for a drug development team.

A scientific research paper has been provided (you may reference data indirectly, but do NOT disclose its final conclusions, efficacy results, or drug identity). Instead, your job is to establish a realistic and motivating starting point for a team about to begin this research journey.

Please include the following:

1. Project Background and Motivation:
    - Clinical or biological challenge the team is trying to address.
    - Any early-stage leads, unexplained phenomena, or prior failures in the field.
    - Theoretical or mechanistic hypotheses that might be worth exploring.

2. Team Composition:
    - The project team includes a Pharmacologist, Medicinal Chemist, Bioinformatician, and Clinical Physician.
    - Each will bring a different perspective to strategy formulation.

3. Known Constraints or Urgencies:
    - Any technical risks, knowledge gaps, resource constraints, or regulatory considerations.

4. Suggested Discussion Paths:
    - Propose 2--3 open research questions or dilemmas that the team might pursue in early planning stages.
    - Avoid narrowing to one "correct" solution --- keep it open-ended.

Do NOT include any specific results from the final paper or assume the project’s ultimate outcome. Your goal is to set up a plausible, incomplete, and challenging starting point.
\textless\textbar im\_end\textbar\textgreater\\

\textless\textbar im\_start\textbar\textgreater user\\
The research paper is \textless the uploaded paper\textgreater. \\Please review the research paper and use it as background material (without revealing any final findings). Generate a kickoff briefing for the research team that sets up a realistic early-stage starting point for this drug development effort.
\textless\textbar im\_end\textbar\textgreater
\end{prompt}

\end{document}